\documentclass[10pt, letterpaper]{article}
\usepackage{cvpr}
\usepackage{times}
\usepackage{epsfig}
\usepackage{graphicx}
\usepackage{amsmath}
\usepackage{amssymb}
\usepackage{enumitem}
\usepackage{url}
\usepackage{mathrsfs}
\usepackage{float}
\usepackage{amsfonts,nicefrac,mathtools,bbm,color}
\usepackage{verbatim}
\usepackage{multicol,lipsum}
\usepackage{titling}

\usepackage[breaklinks=true,bookmarks=false]{hyperref}

\cvprfinalcopy 

\begin{document}

\title{Using Poisson Binomial GLMs to Reveal Voter Preferences}

\author{Evan Rosenman\\
Stanford University\\
{\tt\small rosenman@stanford.edu}
\and
Nitin Viswanathan\\
Stanford University\\
{\tt\small nviswana@stanford.edu}
}

\maketitle



\begin{multicols}{2}

\begin{abstract}
We present a new modeling technique for solving the problem of ecological inference, in which individual-level associations are inferred from labeled data available only at the aggregate level. We model aggregate count data as arising from the Poisson binomial, the distribution of the sum of independent but not identically distributed Bernoulli random variables. We relate individual-level probabilities to individual covariates using both a logistic regression and a neural network. A normal approximation is derived via the Lyapunov Central Limit Theorem, allowing us to efficiently fit these models on large datasets. We apply this technique to the problem of revealing voter preferences in the 2016 presidential election, fitting a model to a sample of over four million voters from the highly contested swing state of Pennsylvania. We validate the model at the precinct level via a holdout set, and at the individual level using weak labels, finding that the model is predictive and it learns intuitively reasonable associations. 

\end{abstract}

\section{Introduction}
\label{Introduction}
For American political parties to compete effectively, it is vital to obtain accurate models of individual voter preferences. Using such models, parties can identify swing voters and focus outreach efforts ahead of elections. 

In the U.S., vote tallies for every candidate in a given race are available at the precinct level, but the most granular data -- who voted for which candidate -- is private. As a result, political groups are forced to rely on polling data to perform analysis at the individual voter level. However, polling data can only provide a representative sample from the population. Moreover, it may be inaccurate, as it was at the state level in the 2016 presidential election \cite{538.polls}. To address this problem, we develop individual voter models based on publicly available precinct-level voting data.

We develop a new type of generalized linear model (GLM), denominated the Poisson binomial generalized linear model, for this task. The model is based on the Poisson binomial distribution, which arises as the distribution of the sum of independent but not identically distributed Bernoulli random variables. To our knowledge this GLM formulation has not previously been implemented in the literature. 

We fit this model to the results from the 2016 presidential election in Pennsylvania, a key swing state that, in an upset, favored Donald Trump over Hillary Clinton by a margin of 0.72\% \cite{NYT}. The model is fit by minimizing the negative log-likelihood of the Poisson binomial, allowing us to discover associations between individual voting preferences and key covariates. 

Our paper makes two key contributions. First, we develop computable approximations to the gradient of the log likelihood, justified by a Central Limit Theorem, allowing us to efficiently train Poisson binomial GLMs. Second, we apply Poisson binomial GLMs to the specific task of revealing voter preferences. 

\section{Related Work}
Theoretical work on the Poisson binomial distribution has focused on computationally tractable ways to estimate its distribution function, often via approximations to other distributions 
\cite{EHM19917, roos1999, chen1974}. Prior research \cite{HONG201341} has identified a closed-form expression for the CDF, which relies on the discrete Fourier Transform. This technique is leveraged in the \texttt{poibin} package \cite{PoibiGithub}, which we use for this paper. The application of the Poisson binomial distribution to the generalized linear model setting has been discussed by Chen and Liu \cite{Poibi}, who propose it for hypothesis testing on the parameter vector for a logistic regression model. 

There is a richer body of literature on modeling voter preferences. Most of this research separates into two primary methodologies. In the first, researchers are interested in the relationship between voter characteristics and ballot preferences. To obtain labeled datasets, researchers use voter surveys \cite{dubrow2007choosing} and exit polls \cite{carsey1995contextual}. They then fit models via simple GLMs, like the multinomial probit or multinomial logit model, which ``represent voter choice as a decision among unordered alternatives (parties, candidates) as a function of chooser (voter) \dots attributes" \cite{DOW2004107}. 

In the second methodology, researchers are interested in the relationship between candidate characteristics or voting methods and vote outcomes. In these studies \cite{miller1998impact, frisina2008ballot, 10.2307/24371899}, researchers frequently use aggregate vote totals from precincts, counties, or states. Relationships are uncovered by linear regression techniques, often using some random effects \cite{10.2307/2585758} and modified to include constraints. Closely related to this literature is the approach of popular election prognosticators like FiveThirtyEight \cite{538} to model election outcomes using a mix of polling and demographic data. 

Lastly, the general problem of ``learning individual-level associations from aggregate data" \cite{flaxman2015supported} has ample precedent in modern machine learning literature \cite{patrini2014almost, kuck2012learning, sun2015message, quadrianto2009estimating}. This is often referred to as the ``ecological inference" problem. Papers from this field typically make weaker assumptions on the functions relating individual-level covariates to aggregate statistics. Kernel methods are widely used, along with probit and logit link functions. Bayesian techniques are then leveraged to actually estimate model coefficients. In the political setting, Flaxman et al. used individual-level covariates to analyze the 2012 \cite{flaxman2015supported} and 2016 \cite{flaxman2016understanding} elections. As will be shown, our approach differs in that is purely frequentist. Our method is less flexible due to its functional form assumptions, but it allows for simpler fitting procedures, straightforward estimation of individual-level probabilities, and greater model interpretability. 

\section{Dataset}

\subsection{Overview}

Two distinct datasets were combined for the purposes of building these models. The dataset of Pennsylvania precinct-level election results contains the total number of votes received by each candidate by precinct in the 2016 presidential election. We obtained this dataset from OpenElections \cite{OpenElections}. The format of the data is represented in Table 1. 
\begin{table*}
\caption{Pennsylvania precinct-level election results}
\begin{center}
\begin{tabular}{ |l|l|l|l| } 
\hline
County Name & Precinct Name & Candidate Name & Number of votes \\
\hline \hline
MONTGOMERY & ABINGTON W1 D1 & HILLARY CLINTON & 603  \\
\hline
MONTGOMERY & ABINGTON W1 D1 & DONALD TRUMP & 388  \\
\hline
... & ...& ... & ...  \\
\hline
\end{tabular}
\end{center}
\end{table*}
\begin{table*}
\label{tab:datasets}
\caption{Pennsylvania voter file}
\begin{center}
\begin{tabular}{ |l|l|l|l|l|l| } 
\hline
County Name & Precinct Name & Voter Name & Gender & Age & Other Attributes \\
\hline \hline
MONTGOMERY & ABINGTON 1-1 & Jane Doe & Female & 27 & ... \\
\hline
... & ... & ... & ... & ... & ... \\
\hline
\end{tabular}
\end{center}
\end{table*}

The other dataset is the Pennsylvania voter file, which we obtained directly from the Pennsylvania Department of State \cite{PAVoterFile}. This dataset contains a row for every registered voter in Pennsylvania as well as their party registration, limited demographic information (age, gender), and voting participation over a set of recent primaries and general elections. The format of the data is represented in Table 2. 

\subsection{Dataset Preparation and Validation}

Because these data were obtained from disparate sources, we ran into challenges in cleanly mapping between the files. In particular, there is no common unique identifier for each precinct, and the precinct names often did not match between the election results and the voter file. As a result, we had to review precincts manually to determine the best way to match them between the files. If either the voter file or precinct-level results were corrupted or did not match with each other, we removed the entire county from our dataset.

After our data cleaning, filtering, and mapping, we ended up with a dataset of 49 of Pennsylvania's 67 counties. Together, these counties represent 7,141 precincts and about 4.37 million total votes. This corresponds to about 71\% of votes cast in Pennsylvania in the 2016 presidential election. The sample had a small pro-Clinton bias, with $51.4\%$ of the voters who cast a major-party ballot supporting Clinton in our sample, versus 49.6\% statewide. 

The Poisson binomial is directly applicable to binary choices, while a more complex model would be needed to account for vote counts among a variety of choices. Hence, to simplify the analysis, we model the election as a binary decision between Hillary Clinton and Donald Trump, excluding third party candidates. 

In practice, this simplifying assumption requires us to take the fraction of \emph{major-party} votes cast in a precinct that went to Clinton, and multiply it by the \emph{total} number of votes in the precinct, to obtain the precinct-level Democratic vote tallies. The analogous process is used to obtain precinct-level Republican vote tallies, and these values are then used to train the model. The process thus ``scales up" the total number of Clinton and Trump votes by a small factor, but it is likely a reasonable approximation, as about 96\% of 2016 Pennsylvania presidential votes went either to Trump or to Clinton. 

\section{Methods}

\subsection{Poisson Binomial GLM}

We use a generalized linear model based on the Poisson binomial distribution. We model an individual $i$ in precinct $k$ voting for Clinton as a Bernoulli random variable with success probability $p_{k, i} = \sigma(\theta^Tx_{k, i})$. Here $\sigma(\cdot)$ denotes the sigmoid function $f(z) = \frac{1}{1 + e^{-z}}$, $\theta$ is a parameter vector (shared across precincts) to fit, and $x_{k, i}$ are known covariates for voter $i$ from the Pennsylvania voter file, as well as an intercept term. The probability of voting for Trump is $1 - p_{k, i}$. We make the additional simplifying assumption that these Bernoulli random variables, representing individual voter choices, are independent. 

Under our stated assumptions, the total number of Clinton voters in each precinct will be a Poisson binomial random variable, the sum of independent but not identically distributed Bernoulli random variables \cite{Poibi}. For a precinct $k$ with $D_k$ votes for Clinton out of $T_k$ total votes, the log likelihood is given by:
\[ \ell_k(\theta) = \log \left( \sum_{A \in F_k} \prod_{i \in A} p_{k,i} \prod_{j \in A^c}(1 - p_{k, j}) \right), \]
where $F_k$ is the set of all configurations of $T_k$ votes in which a total of $D_k$ votes were cast for Clinton; $A$ is the set of voters who voted for Clinton under that configuration, and $A^c$ is the set of voters who voted for Trump under that configuration. The log likelihood of the precinct-level results given parameters $\theta$ can be calculated by adding the likelihoods from every precinct together:

\[\ell(\theta) = \sum_k \ell_k(\theta) \,.\]

In order to determine the optimal parameters $\theta$, we need to maximize the likelihood over the Pennsylvania precincts. Note that the Poisson binomial likelihood involves sums over all possible configurations of votes -- e.g. if Clinton received 200 out of 500 total votes in a precinct, then the likelihood involves a sum over $500 \choose 200$ configurations. Although we can directly estimate the likelihood using the \texttt{poibin} package, calculating the gradient is computationally infeasible.

\subsection{Calculating the Gradient}

To address this problem, we make use of the Lyapunov CLT \cite{billingsley1995probability} to observe that the asymptotic distribution of $D_k$ is given by: 
\[ D_k \stackrel{d} \longrightarrow N \left(\sum_{i} p_{k,i}, \sum_{i} p_{k, i}(1-p_{k, i}) \right) \,. \] 
This result is proven in the appendix. It allows us to \emph{estimate} the log likelihood with a much simpler function of $\theta$. In this case, the contribution of precinct $k$ to the overall log likelihood is approximately
\[ \ell_k(\theta) \approx -\log \left( \phi_k\right) + \frac{1}{\phi_k^2} \left( D_k -\mu_k \right)^2,\]
where irrelevant constants have been dropped, $\mu_k =  \sum_{i} p_{k, i}, \phi_k^2 = \sum_{i} p_{k, i}(1-p_{k, i})$, and $p_{k, i} = \sigma(\theta^T x_{k, i})$. This yields a gradient of the form: 
\begin{align*}
&\nabla_{\theta} \ell_k \approx \frac{1}{\phi^2} (D_k - \mu_k) \left( \sum_i p_{k, i} (1 - p_{k, i}) x_{k, i} \right)-\\&  \frac{1}{2} \left(\frac{(D_k - \mu_k)^2}{\phi^4} - \frac{1}{\phi^2} \right) \left( \sum_{i} (2 p_{k, i} - 1)(1-p_{k, i})p_{k, i}x_{k, i} \right) \,.
 \end{align*}

As mentioned, our approach differs substantially from the approaches taken in the related literature. Specifically, we focus on a generalized linear model relating the individual covariates to Bernoulli random variables, and make an additional independence assumption in order to obtain a Poisson binomial distribution for the $D_k$. This model allows us to use a Central Limit Theorem to approximate the likelihood, which allows us to efficiently estimate model parameters. 

\subsection{Neural Network} 
As one potential improvement, we investigated using a neural network rather than a logistic regression to relate individual-level features to the probability of voting for Clinton. In particular, our new model for $p_{k, i}$, the Clinton-voting probability for person $i$ in precinct $k$, is given by: 
\begin{align*}
h_{k, i} &= \sigma\left(W_1 x_{k, i} + b_1 \right) \\
p_{k, i} &= \sigma\left(W_2 h_{k, i} + b_2 \right) \, ,
\end{align*}
where $W_1, W_2$ are weight matrices and $b_1, b_2$ bias vectors. We can compute the gradients: 
\begin{align*}
\frac{\partial \ell_k}{\partial p_{k, j}} &=  \frac{-1 + 2p_{k, j}}{2 \phi_k^2} + \frac{1 - 2p_{k, j}}{2 \phi_k^4} (D_k - \mu_k)^2  + \frac{D_k - \mu_k}{\phi_k^2}\\
\frac{\partial \ell_k}{\partial b_2} &= \sum_{j = 1}^{n_k} \frac{\partial \ell_k}{\partial p_{k, j}} p_{k, j} (1 - p_{k, j}) \\
\frac{\partial \ell_k}{\partial W_2} &= \left\{ h_{k, j} \right\}_j^T \left\{ \frac{\partial \ell_k}{\partial p_{k, j}} p_{k, j} (1 - p_{k, j})\right\}_{j} \\
\frac{\partial \ell_k}{\partial b_1} &= \sum_{j = 1}^{n_k} \left\{ \frac{\partial \ell_k}{\partial p_{k, j}} p_{k, j} (1 - p_{k, j})\right\}_{j} W_2^T \circ  \\ & \hspace{5mm}  \left\{ h_{k, j}(1 - h_{k, j})\right\}_j \\
\frac{\partial\ell_k}{\partial W_1} &= \left\{ x_{k, j} \right\}_j^T \left\{ \frac{\partial \ell_k}{\partial p_{k, j}} \cdot p_{k, j} (1 - p_{k, j})\right\}_{j} W_2^T \circ \\ & \hspace{5mm} \left\{ h_{k, j}(1 - h_{k, j})\right\}_j \,,
\end{align*}
where $\{x_j\}_j$ denotes a column vector consisting of the entries $x_j$ and $\circ$ denotes a Hadamard product.

For our experiments, we construct a feedforward neural network with one hidden layer with 10 neurons. We experimented with other sizes for the hidden layer but saw little change in performance.

\subsection{Optimization Methodology}

Our goal is to obtain a good estimate of the probability of any given voter supporting Clinton. We train our model using gradient descent on the negative log likelihood, where the log likelihood and gradients are as defined in sections 4.2 and 4.3. For each epoch, we train our model on every precinct in the training set, looping through them in the same order every time and updating parameter values after each precinct is encountered. We explored using batch and stochastic gradient descent, but both of these resulted in slower convergence. We used a learning rate of 0.0001 with annealing at a $n^{-1/2}$ decay. 

In the logistic regression case, we initialize our coefficients to zero (yielding a 50\% Clinton-voting probability for every voter). For the neural network, we initialize $W_1, W_2$ to small random uniform values in $[-0.1, 0.1]$ and the biases to zero. 

One specific issue we encountered when running gradient descent was extremely large or small gradients. Small gradients were simply ignored in the training loop, as they would not have altered the parameters meaningfully. Large gradients were clipped and scaled to a fixed $L_2$ norm, allowing a substantive but not enormous parameter update in the desired direction. 

We evaluated the fit of our model in a number of ways. First, we ensured that the loss (i.e. negative log-likelihood) decreased over time. Decreasing loss over training epochs indicates that our model is improving, and we look for it to eventually stabilize, which indicates that the parameter values have approximately converged.

\section{Results \& Analysis}

\subsection{Evaluating Model Accuracy}
We began by validating that the models were training effectively, and were accurate in predicting voter preferences in aggregate. We split the dataset into a training subset, consisting of 70\% of the precincts, and a test subset, consisting of the remaining 30\%. We then trained the model, computing the Poisson binomial negative log likelihood over the entire dataset, and the Gaussian approximation, after each epoch. 
\begin{figure}[H]
\vskip 0.2in
\begin{center}
\centerline{\includegraphics[width=\columnwidth]{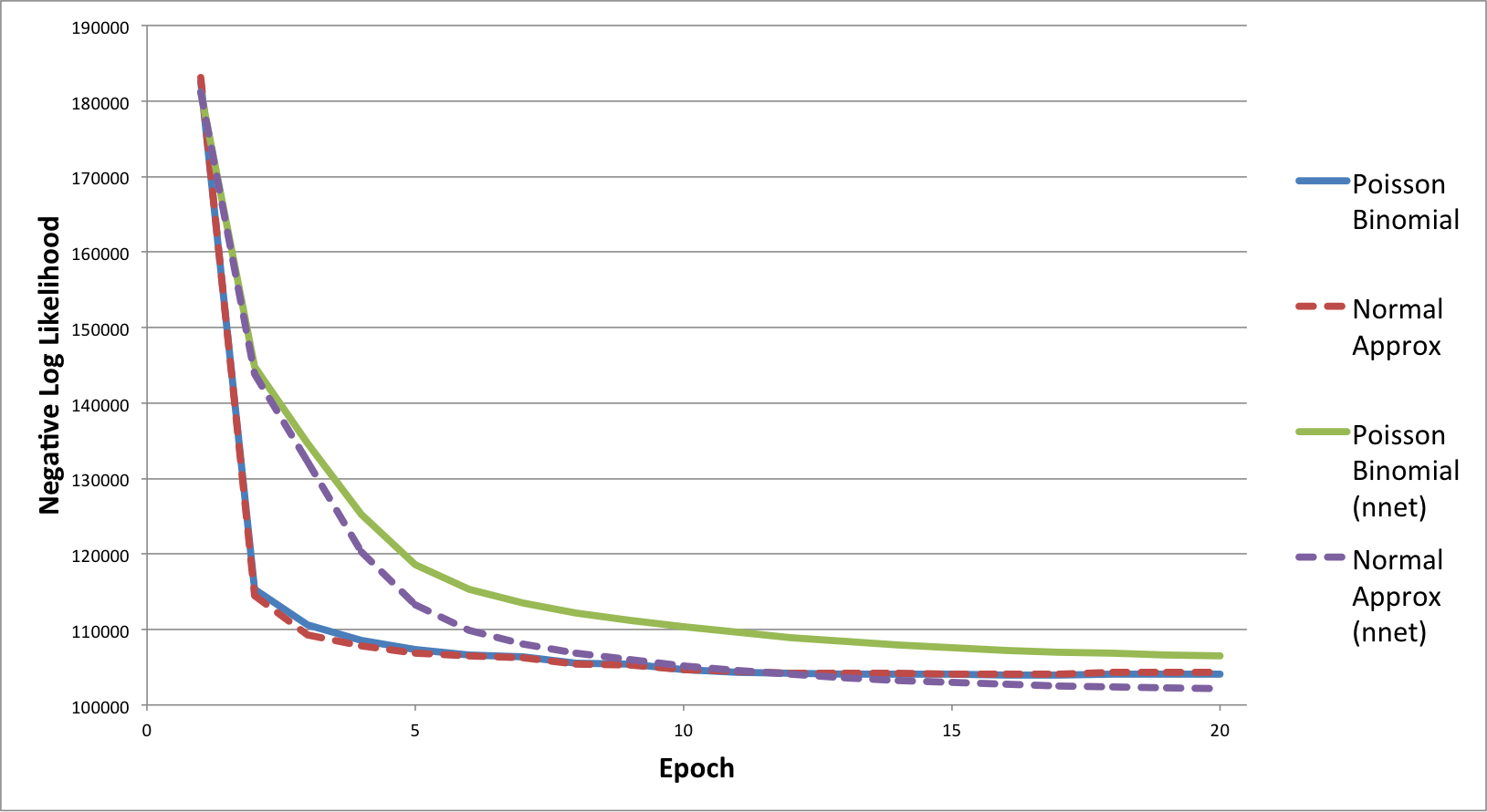}}
\caption{True and approximate training loss (negative log likelihood) over 20 epochs for logistic regression and neural network models}
\label{fig:training-loss}
\end{center}
\end{figure}

We can see in Figure \ref{fig:training-loss} that the Normal approximation is generally a reasonable proxy for the Poisson binomial loss. The logistic regression model trains very quickly, with the training loss diminishing rapidly in the first few epochs, and limited benefits after about 10 epochs. The neural network trains more slowly -- which is consistent with the need to fit many more parameters -- but also appears to approximately converge in 20 epochs. 

When the models were evaluated on a test set, they exhibited somewhat divergent behavior based on how the training and test sets were defined. In the first case, we again defined the 70\%-30\% split at the precinct level, meaning every county had some precincts in training and some in test. 

\begin{figure}[H]
\vskip 0.2in
\begin{center}
\centerline{\includegraphics[width=\columnwidth]{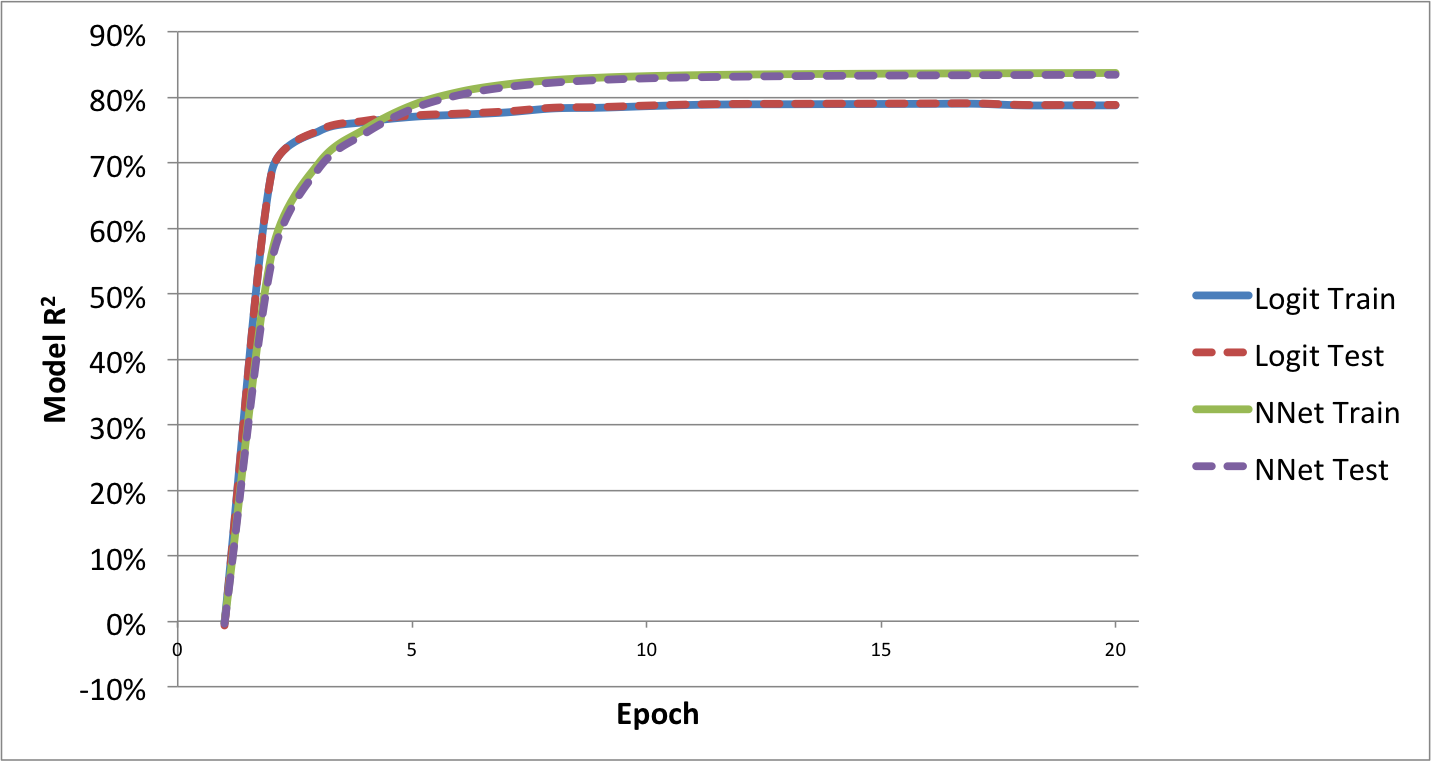}}
\caption{Training and test $R^2$ for logistic regression and neural network models, with sets split at precinct level}
\label{fig:holdout-R2-precinct}
\end{center}
\end{figure}

Figure \ref{fig:holdout-R2-precinct} shows model $R^2$ when predicting vote proportions within each precinct, for both the training and holdout sets. Results are weighted by the vote count in the precinct. In this case, both models exhibited virtually identical behavior in training and test, indicating no overfit. The neural network model trained more slowly, but ultimately outperformed with a final test $R^2$ of 83.5\% vs. 78.9\% for the logistic regression model.

\begin{figure}[H]
\vskip 0.2in
\begin{center}
\centerline{\includegraphics[width=\columnwidth]{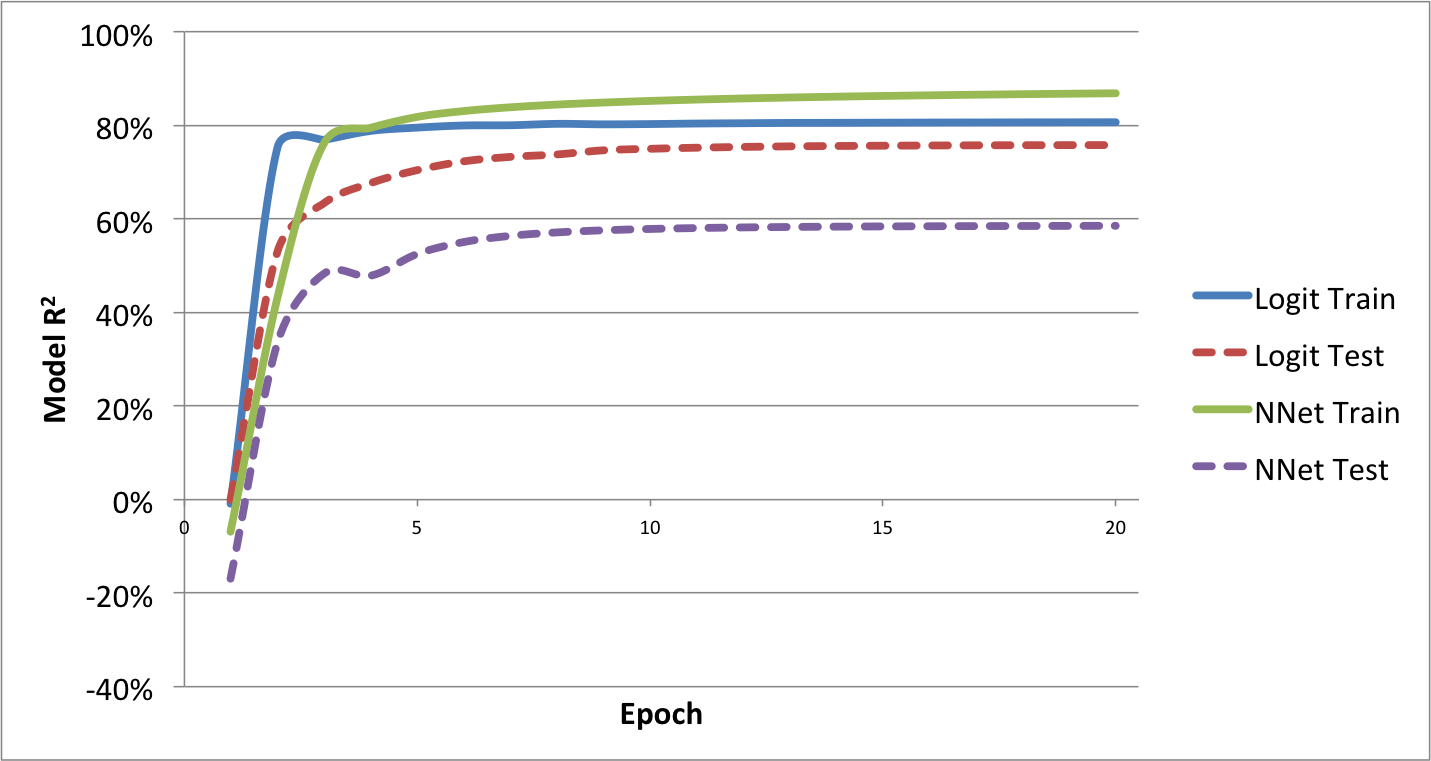}}
\caption{Training and test $R^2$ for logistic regression and neural network models, with sets split at county level}
\label{fig:holdout-R2-county}
\end{center}
\end{figure}

In Figure \ref{fig:holdout-R2-county}, the 70\%-30\% split was implemented at the \emph{county} level, such that every county was either entirely in the training set or entirely in the test set. This induces greater dissimilarity between training and test, though the split was chosen such that the overall Clinton vote share was very similar (within 2.5\%) between training and test. 

Now, we see the opposite behavior: training $R^2$ is higher than test $R^2$ for both models, but the gap is much larger for the neural network. As a result, the final test $R^2$ of 75.8\% for the logistic regression model was quite a bit higher than the 58.6\% achieved by the neural network model. In this case, the neural network appears to be fitting ``too hard" to the training counties, and thus its performance on the test counties degrades considerably. 

We also tried adding $L_2$ regularization to the neural network model in order to reduce overfit. But across a wide range of $\lambda$ values, the test set accuracy consistently remained at or below 58\%, indicating little benefit. It is possible that more complex regularization schemes, such as dropout, might induce better performance on the test counties. 

The case of splitting training and test sets by county is likely a more realistic proxy for how this model would be used in practice. For example, models might be fit to one geographic area or one group of voters and then applied to another similar but non-identical population. Hence, the robustness (and interpretability) of the logistic regression model is a significant advantage. We hence focus on this model and the county-level train vs. test set splitting scheme for the remainder of the analysis. 





\subsection{Weak Label Results}

We next sought to validate model accuracy at the individual voter level. However, due to the nature of voting data and the secret ballot, we do not actually have any labeled data that we can use to evaluate our model. To address this issue, we used weak labels as an approximate way to gauge our model's performance. Weak labels -- ``noisier, lower-quality" labels that are imperfect but easily obtainable -- are a common technique used in supervised learning problems where either no labeled data is available or the available data is insufficient \cite{rbvr2017}. 

We applied two different forms of weak labels to gauge model performance at the individual voter level. First, we evaluated our model only against landslide precincts in the test set, where we define a landslide precinct to be one in which 90\% or more of voters supported the same candidate. We used our model to generate predictions for each voter in these precincts. We expected to see very high predicted probabilities of voting for Clinton among voters in Democratic landslide precincts, and the converse for voters in precincts that went to Trump. 

\begin{figure}[H]
\vskip 0.2in
\begin{center}
\centerline{\includegraphics[width=\columnwidth]{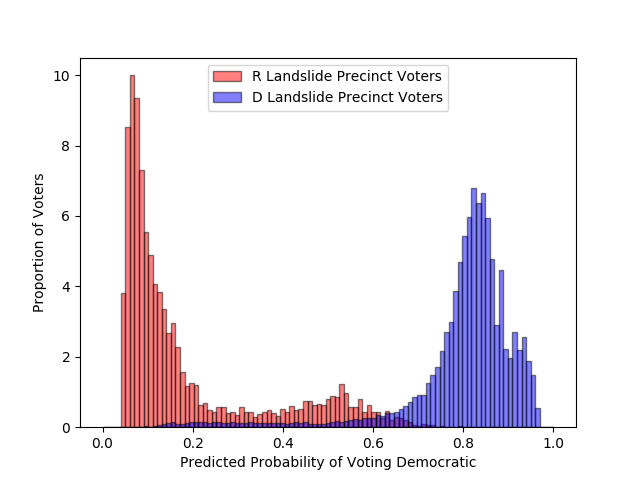}}
\caption{Predictions for voters in landslide precincts}
\label{fig:lanslideHist19}
\end{center}
\end{figure}

As expected, Figure \ref{fig:lanslideHist19} has a bimodal distribution: the vast majority of voters are either extremely likely to vote for Clinton or extremely likely to vote for Trump. The average among voters in Trump landslide precincts is an 18.9\% probability of voting for Clinton, and the average among voters in Clinton landslide precincts is 79.6\%, which does represent some shrinkage toward the mean from the true probabilities, which lie in $[0\%, 10\%]$ and $[90\%, 100\%]$ respectively. Nonetheless the populations are generally well-separated, as desired. 

In our second weak labeling validation, we assumed that all primary voters supported their party's eventual nominee in the general election. This is, of course, an approximation -- one well-regarded study found that about 12\% of each party's primary voters supported the opposing party's candidate in the 2016 general election \cite{GDF6Z0_2017}. To evaluate, we built our model on the training counties, then evaluated our model on all voters in the test counties who voted in a primary.

\begin{figure}[H]
\vskip 0.2in
\begin{center}
\centerline{\includegraphics[width=\columnwidth]{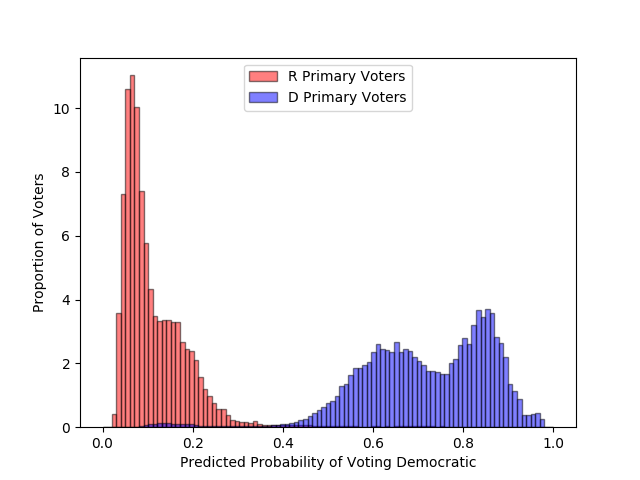}}
\caption{Predictions for primary election voters}
\label{fig:primaryVoters19}
\end{center}
\end{figure}

Figure \ref{fig:primaryVoters19} demonstrates the desired distributional separation. We again see some shrinkage toward the middle, particularly among the Democratic primary voters, who are assigned an average of 72\% probability to vote for Clinton in the general election, versus 11.6\% for Republican primary voters. 

\subsection{Final Model Trained on All Data}
After confirming the model's predictive validity, we trained on our entire dataset to obtain final coefficients, which are provided in the table below. Positive values correspond to a higher probability of voting for Clinton. Note that not all individuals have a gender listed in the voter file.  

\begin{table}[H]
\caption{Model coefficients. Positive values correspond to a higher probability of voting for Clinton. The Age variable provided to the model is Age/40, to put it on a similar scale to other variables}
\begin{tabular}{|l|c|}
\hline
Model Term & Value \\
\hline 
Apartment Dweller & 1.246 \\
Registered Democrat & 1.028  \\
Voted in the Democratic Primary & 0.268  \\
County College Educated \% & 0.256  \\
County Population Density & 0.197  \\
Voted Absentee 2012 & 0.157 \\
is Female & 0.097  \\
Voted Absentee 2014 & 0.086 \\
County Latino \% & 0.075  \\
County Black \% & 0.040  \\
Voted In Person 2014 & -0.083 \\ 
County Income & -0.093  \\
County White \% & -0.158  \\
Voted In Person 2012 & -0.183 \\
is Male & -0.267\\
Voted in the Republican Primary & -0.507  \\
Age* & -0.562  \\
Registered Republican & -1.328  \\
\hline
\end{tabular}
\label{table:coeff}
\end{table}

The signs of these coefficients are all in line with our expectations: Democrats, younger voters, urban voters, and women are more likely to vote for Clinton while the opposite is true for Trump. 

The relative magnitudes provide a more nuanced picture. Notably, the model finds a stronger association between being a registered Republican and voting for Trump than being a registered Democrat and voting for Clinton. This makes sense, given that our sample of voters was 49.4\% registered Democrats, 39.7\% registered Republicans, and 10.8\% registered with neither party. Though independents likely broke disproportionately for Trump, the nearly 10\% Democratic registration advantage among voters in our sample -- as compared to the virtually equal vote tallies -- implies that registered Democrats were less likely to vote for Clinton than registered Republicans were to vote for Trump. 

This discrepancy also partly explains why being an apartment dweller -- rather than being a registered Democrat -- is the strongest predictor of voting for Clinton. Clinton won Philadelphia county -- home to Pennsylvania's largest city -- with over 84\% of the vote  \cite{NYT}. She came nowhere close to that margin in any other county in the state, while many non-urban counties with large Democratic voter registration advantages nonetheless voted for Trump  \cite{NYT}. Thus, urbanity indeed seems to be a strong individual-level predictor of Democratic voter preference in Pennsylvania. 

\section*{Conclusions and Future Work}

We have combined aggregated precinct-level election data with information about individual voters to build a model predicting how individuals voted. We implemented both logistic regression and neural network models to relate individual covariates to voting probabilities. We then trained these models via gradient descent, making use of a normal approximation to the Poisson binomial, which we derived. Our model's accuracy was validated using both aggregate vote totals and two weakly labeled test sets at the individual level. After validating model accuracy, we retrained the model on the entire dataset to obtain our final model for individual voter preferences in Pennsylvania during the 2016 presidential election. We believe this model would be a valuable resource for political organizations in voter targeting. 

Our work contributes to the literature on learning individual associations from aggregate data, and is one of only a small number of such papers to work directly on political data. Our fitting method appears to be novel, as we find no references in the literature to leveraging the Poisson binomial distribution in this way. 

For future work, we'd like to further explore the idea of using neural networks to compute individual voter probabilities. With further tuning of hyperparameters, we might obtain a neural network model that outperforms the logistic regression model while being less susceptible to overfit. We could also look at other regularization schemes, such as dropout, to reduce the gap between the training and test performance. We are also curious to explore probit and cauchit models to predict individual voter behavior from covariates. 

Lastly, we would like to develop data pipelines that would allow us to efficiently repeat this analysis on other states and elections throughout the U.S. 

\section*{Acknowledgements}

Evan Rosenman is supported by the Department of Defense (DoD) through the National Defense Science \& Engineering Graduate Fellowship (NDSEG) Program. We also thank Art Owen and Michael Baiocchi for their guidance and feedback on this work. 

\newpage

{
\bibliographystyle{ieee}
\bibliography{csbib}
}

\end{multicols}

\newpage
\section*{Appendix: Lyapunov CLT Proof}

Define $D_k = \sum_{i = 1}^n D_{k, i}$ to be the number of Democratic votes in precinct $k$, where $D_{k, i}$ is an indicator variable denoting whether person $i$ in precinct $k$ voted for Clinton. $D_k$ follows a Poisson binomial distribution with success probabilities $p_k = (p_{k, 1}, \dots, p_{k, n})$. Define $s_{k}^2 = \sum_{i = 1}^{n} p_{k, i} (1-p_{k, i})$. 
We check the Lyapunov CLT \cite{billingsley1995probability} condition for the fourth moment: 

\begin{align*}
\lim_{n \to \infty} & \frac{1}{s_k^4} \sum_{i = 1}^n E \left( (D_{k, i} - p_{k, i})^4 \right) = \\ & \lim_{n \to \infty} \frac{\sum_{i = 1}^n p_{k, i}(1-p_{k, i}) \left( 3p_{k, i}^2 - 3p_{k, i} + 1 \right)}{\left(\sum_{i = 1}^n p_{k, i} \left( 1- p_{k, i} \right) \right)^2} \stackrel{?} = 0 
\end{align*}

Observe that $3p_{k, i}^2 - 3p_{k, i} + 1 \in (0, 1)$ if $p_{k, i} \in (0, 1)$. Hence, the numerator is strictly less than $\sum_{i = 1}^n p_{k, i} (1 - p_{k, i})$. Thus, if we can guarantee the numerator grows without bound, then this limit is 0 and the Lyapunov CLT applies. We can do so using a simple condition, like enforcing that there is some $\epsilon > 0$ such that $\epsilon < \bar p_i < 1- \epsilon$ for all $i$ (i.e. the mean probability of voting for Clinton in a precinct never falls below some infinitesimal threshold $\epsilon$ or above $1-\epsilon$). 

The Lypaunov CLT now tells us that: 
\[ \frac{D_k - \sum_{i = 1}^{n} p_{k, i} }{s_k} \stackrel{d} \longrightarrow N(0, 1) \] 
giving us the desired asymptotic normality.

\end{document}